%% file: main_1.tex
\documentclass[conference, a4paper]{IEEEtran}
\usepackage{cite}
\usepackage{amsmath,amssymb,amsfonts}
\usepackage{algorithmic}
\usepackage{graphicx}
\usepackage{textcomp}
\usepackage{hyperref}
\usepackage{xcolor}
\usepackage{booktabs}
\usepackage{float}
\def\BibTeX{{\rm B\kern-.05em{\sc i\kern-.025em b}\kern-.08em
    T\kern-.1667em\lower.7ex\hbox{E}\kern-.125emX}}
    
\newcommand{\G}{G}
\newcommand{\dataset}{X}
\newcommand{\pdatadistw}{p_\dataset}
\newcommand{\pdatadist}[1]{\pdatadistw(#1)}
\newcommand{\pGdatadistw}{\Tilde{p}_\dataset}

\newcommand{\xtarget}{\dot{x}}
\newcommand{\x}{x}
\newcommand{\traingset}{X}
\newcommand{\z}{z}
\newcommand{\Z}{Z}
\newcommand{\xhat}{\hat{\x}}
\newcommand{\zhat}{\hat{z}}
\newcommand{\zhatbase}{z_0}
\newcommand{\loss}{L}
\newcommand{\zprior}{\Z}
\newcommand{\zpriordisw}{p_\Z}
\newcommand{\zpriordis}[1]{\zpriordisw(#1)}
\newcommand{\distfun}{d}
\newcommand{\zspace}{\mathcal{Z}}
\newcommand{\Xspace}{\mathcal{X}}
\newcommand{\zreg}[1]{\mu}

\newcommand{\cn}[4]{\textit{#1}-\textit{#2}-\textit{#3}-$#4$}
\newcommand{\odlv}{\textit{OLV} }

\begin{document}

\title{Adversarial Out-domain Examples for Generative Models}
\author{\IEEEauthorblockN{1\textsuperscript{st} Dario Pasquini}
\IEEEauthorblockA{\textit{Department of Computer Science} \\
\textit{Sapienza University}\\
Rome, Italy \\
pasquini@di.uniroma1.it}
\and
\IEEEauthorblockN{2\textsuperscript{nd} Marco Mingione}
\IEEEauthorblockA{\textit{Department of Statistics} \\
\textit{Sapienza University}\\
Rome, Italy \\
marco.mingione@uniroma1.it}
\and
\IEEEauthorblockN{3\textsuperscript{rd} Massimo Bernaschi}
\IEEEauthorblockA{\textit{Institute for Applied Computing (IAC)} \\
\textit{CNR}\\
Rome, Italy \\
massimo.bernaschi@cnr.it}
}

\maketitle

\begin{abstract}
Deep generative models are rapidly becoming a common tool for researchers and developers. 
However, as exhaustively shown for the family of discriminative models, the test-time inference of deep neural networks cannot be fully controlled and erroneous behaviors can be induced by an attacker. In the present work, we show how a malicious user can force a pre-trained generator to reproduce arbitrary data instances by feeding it suitable adversarial inputs. Moreover, we show that these adversarial latent vectors can be shaped so as to be statistically indistinguishable from the set of genuine inputs.
The proposed attack technique is evaluated with respect to various \textit{GAN} images generators using different architectures, training processes and for both conditional and not-conditional setups.
\end{abstract}
%
%
\section{Introduction}
The existence of adversarial inputs has been demonstrated for a quite large set of deep learning architectures \cite{adv, advseg, advrl}. An adversarial input is a carefully forged data instance that aims at leading the model to behave in an incorrect or unexpected way. Moreover, the adversarial setup requires that those instances must be indistinguishable from genuine inputs.

In the present work, motivated by the extensive studies carried out on adversarial inputs for discriminative models, we extend the adversarial context into the increasingly popular generative models field. In particular, we focus on the most promising class of architectures, called \emph{Generative Adversarial Networks} (\textit{GANs}) \cite{goodfellow2014generative}. \textit{GANs} perform generative modeling of a target data distribution by training a deep neural network architecture. This is composed by two neural networks, a generator and a discriminator that are trained simultaneously in a zero-sum game. In the end, the generator learns a deterministic mapping between a latent representation and an approximation of the target data distribution.
\begin{figure}[ht]
\centering
\includegraphics[scale=0.30]{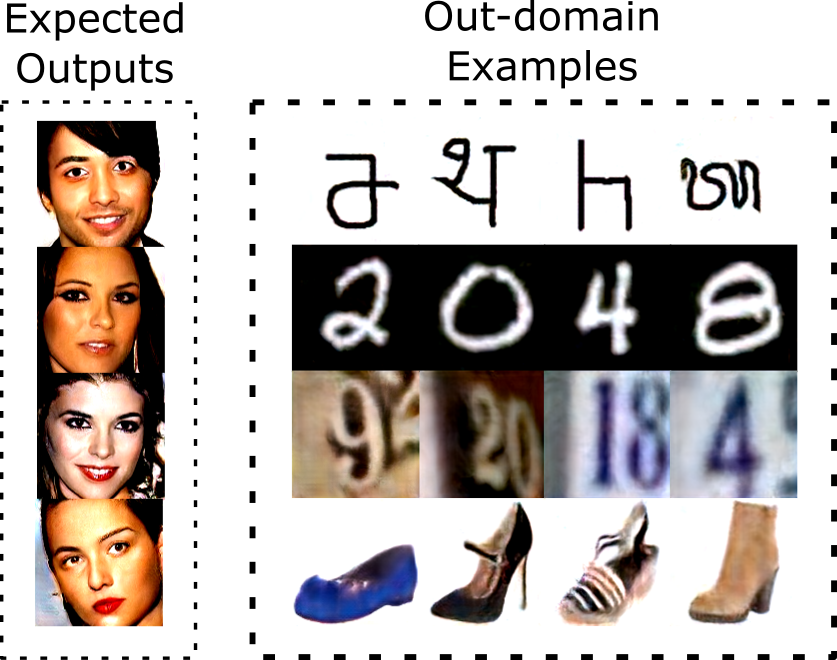}
\caption{\small Comparison between expected generator outputs (left column) and generated out-domain examples (right column) for a Progressive \textit{GAN} generator trained on the \textit{CelebA} dataset.}
\label{progan_odlv}
\end{figure}
What we show with the present work is that a pre-trained generator can be forced to reproduce an arbitrary output if fed by a suitable adversarial input. In particular, our findings show that the data space, defined by the generator, contains data instances having very low probability of lying in the space of the expected outputs (i.e., the target data distribution). We will refer to those outputs as \textit{out-domain examples} and to the relative adversarial inputs as \textit{out-domain latent vectors} or \odlv in short.
Figure \ref{progan_odlv} shows a set of \textit{out-domain examples} for a generator trained by using a Progressive GAN \cite{progan}. In that example, we found a set of inputs capable to force the generative model\footnote{ A pre-trained \textit{Progressive GAN} available at \url{https://tfhub.dev/google/progan-128/1}.} to produce images completely different from those belonging to the generator training dataset, i.e., the \textit{CelebA} dataset \cite{celeba}. Moreover, we show that those \odlv can be forged in order to be \emph{statistically} indistinguishable from known-to-be trusted inputs. The existence of such adversarial inputs raises new practical questions about the use of \textit{GAN} generators in an untrustworthy environment, as a web application. The main contributions of the present paper may be summarized as follows:
\begin{enumerate}
    \item We show that a generator may be forced to produce \emph{out-domain} data instances which are arbitrarily different from those for which the generator is trained. 
    Our experiments refer to three common image datasets and for standard and conditional \textit{GAN} architectures: \emph{Deep Convolutional GAN} (\textit{DCGAN}) \cite{dcgan} and \emph{Auxiliary Classifier GAN} (\textit{ACGAN}) \cite{acgan}.
    \item We propose a first type of adversarial input for not encoder-based generative models. 
    \item We investigate the nature of out-domain examples showing that their quality strongly depends on the dimension of both the latent and the data space.
\end{enumerate}

%
%
%
\section{Background}
\label{bk}
The objective of a generative model is to learn a probability distribution $\pGdatadistw$ that approximates a target data probability distribution $\pdatadistw$. Actually, in general, $\pdatadistw$ is unknown and it can only be inferred by a limited set of samples.
One of the most powerful approaches to train a generative model is the recently proposed \emph{Generative Adversarial Networks} (\textit{GANs}) framework. \textit{GANs} require the simultaneous training of two neural networks: a generator $G$ and a discriminator $D$. $D$'s objective is to maximize the probability of discriminating between $\pGdatadistw$ and $\pdatadistw$, whereas $G$'s objective is to make $\pGdatadistw$ and $\pdatadistw$ indistinguishable in order to mislead $D$. This kind of unsupervised training process is renamed \emph{Adversarial Training} in this context. \\
From the mathematical point of view, both the generator and the discriminator can be intended as functions: $G: \zspace \rightarrow \Xspace$ and $D: \Xspace \rightarrow [0, 1]$. 
In other words, during the training process, the neural network $G$ receives as input a vector $z = (z_1, \dots, z_n)$ and produces $\x = \G(\z)$. Each element of $z$ is a realization of a random vector $Z_1, \dots, Z_n$, with $Z_i \overset{i.i.d.}{\sim} p_Z$, where $p_Z$ is an arbitrary density function.
Then, the optimization problem can be easily summarized by the following formulation:
\begin{equation}
    \label{ganloss}
    \min_{\theta_G}\max_{\theta_D}\langle \mathbb{E} [\log(D(x))] + \mathbb{E}[ \log( D(1-G(z)))] \rangle
\end{equation}
where $\theta_G$ and $\theta_D$ are, respectively, the parameters of the generator and the discriminator network and $x$ is an instance from the training set $\traingset$.\\ 
%
%
The use of random latent instances as input of $\G$ makes possible to explore the data space by generating new data instances, not necessarily available in the training set.\\
Many extensions to the original \textit{GAN} framework have been successfully developed \cite{dcgan,acgan,wgan,donahue2016adversarial}. One of the most influential work is \cite{dcgan}, in which the \textit{DCGAN} architecture was proposed. \textit{DCGAN} is capable of exploiting the potential of \emph{Convolutional Neural Networks} \textit{(CNNs)} in both the generator and  discriminator perspective.
The \textit{GAN} framework can be easily extended to train conditional generators \cite{mirza2014conditional} using the \textit{ACGAN} architecture \cite{acgan}. In this case, a supervised training approach is used to the purpose of learning a probability distribution conditioned to a set of classes $Y$. During this training process, the class $y$ is chosen randomly (e.g., with uniform probability) from the set $Y$ including all the possible classes. Then, the generator is modeled as a bivariate function that receives as input an instance of the latent space $z$ labelled with its related class $y \in Y$. \textit{ACGAN} architecture improves the performance of the training process by adding an auxiliary classification task to the discriminator. The latter outputs two probability distributions, the first over the source (i.e., the probability that the instance comes from $\pdatadistw$) and the second over the class labels (i.e., the probability that the instance belongs to class $y$). 
In this case, the optimization problem can be parametrized by extending \eqref{ganloss} as follows: 
\begin{align*}
L_{\text{source}} = \mathbb{E}[ \log( D^{\text{source}}(x))]\ +\ 
\mathbb{E}[ \log( D^{\text{source}}(1-G(z)))]
\end{align*}
\begin{align*}
L_{\text{class}} = \mathbb{E}[ \log( D^{\text{class}}(y | x))]\ +\ 
\mathbb{E}[ \log( D^{\text{class}}(y | G(z, y)))]
\end{align*}
\begin{equation}
L_{D} = L_{\text{class}}+L_{\text{source}}\qquad 
L_{G} = L_{\text{class}}-L_{\text{source}}
\end{equation}
where $L_{D}$ and $L_{G}$ are the loss functions of the discriminator and the generator, respectively.
\section{Related works}
\label{rel_works}
Some examples of adversarial inputs for encoder-based generative models like \emph{Variational AutoEncoder} (\textit{VAE}) and \textit{VAE-GAN} are analyzed in \cite{vae_adversarial}. In the proposed scenario, given a data instance $x$, an attacker aims at producing an instance $\hat{x}$ that differs in a limited way from $x$ but which is capable of driving the \textit{VAE} to reconstruct a $\hat{x}$ far from the original $x$. The reconstructed $\hat{x}$ can be an approximation of an arbitrary data instance chosen from the attacker. All the results described by the authors refer to $\hat{x}$ as if it came from the same data distribution on which the \textit{VAE} is trained. A similar attack scenario has been investigated also in \cite{vae_attack2}. At the best of our knowledge no other form of adversarial input targeting \textit{GAN} generators has been proposed.\\
Many works investigate the possibility of finding or exploiting an inverse mapping from the data space to the latent space of a pre-trained generator \cite{nguyen2017plug, creswell2018inverting, precise_recovery}.
In \cite{nguyen2017plug} a pre-trained generator $G$ is used to invert a discriminative model $C$ to the purpose of synthesizing novel images. It is noteworthy that the authors report a first case of partial out-domain example. In particular, a generator trained on \textit{ImageNet} was able to reproduce images belonging to classes known to $C$ but unknown to $G$.\\
Additionally, a technique to map images into a latent representation with a pre-trained generator is proposed in \cite{creswell2018inverting} and in \cite{precise_recovery}. In those works, the authors mention the possibility of mapping images, which are not present in the training set, into the generator latent space. That is shown only for images coming from the same distribution on which the model is trained. The proposed inversion technique is essentially the same used in the present work and it is based on the direct optimization of the generator input by  a gradient descent based approach. In \cite{creswell2018inverting} during the optimization process, the latent vectors are encouraged to be \emph{similar} to those of the latent prior distribution by adding a penalty term in the loss function. This term is a weighted sum of the discrepancy between the mean and standard deviation of the latent vector and the latent prior distribution. We exploited the possibility of extending this penalty term beyond the second moment, given that just two moments might not be sufficient to correctly identify the latent vectors.\\
The same generator inversion technique is used in \cite{DefenseGAN} as a defense against adversarial examples. In this work, the adversarial examples are mapped to unperturbed data instances by inverting a generator trained on the data distribution of clear data. The proposed model inversion aims at finding the closest generator codomain element for each input of a discriminative model $f$. At the end, these codomain elements are taken as input by $f$ instead of the original untrusted data.\\
A similar technique is also used in \cite{membGAN} to perform a data membership attack against generative models. That kind of attack aims at inferring the presence of a data instance $x$ in the training-set used during the training of a generative model $G$. In this case, the generator inversion is carried out by a neural network attacker called $\mathcal{A}$. This attacker is trained as an encoder for $G$. Given a data instance $x$, the latent vector $z = \mathcal{A}(x)$ is used to estimate the chances of $x$ of being in the $G$'s dataset by calculating the distance between $G(z)$ and $x$.
\section{Out-domain examples}
\label{ode}
Let $\Xspace$ be the set of all possible data that can be generated by $\G$ and let $\zspace$ be the set of all the possible latent vectors coming from $\zpriordisw$. Then, an out-domain example for a generator $\G$ is defined to be an element $\xhat \in \Xspace$ such that:
\begin{figure*}[ht]
\centering
\includegraphics[scale=0.25]{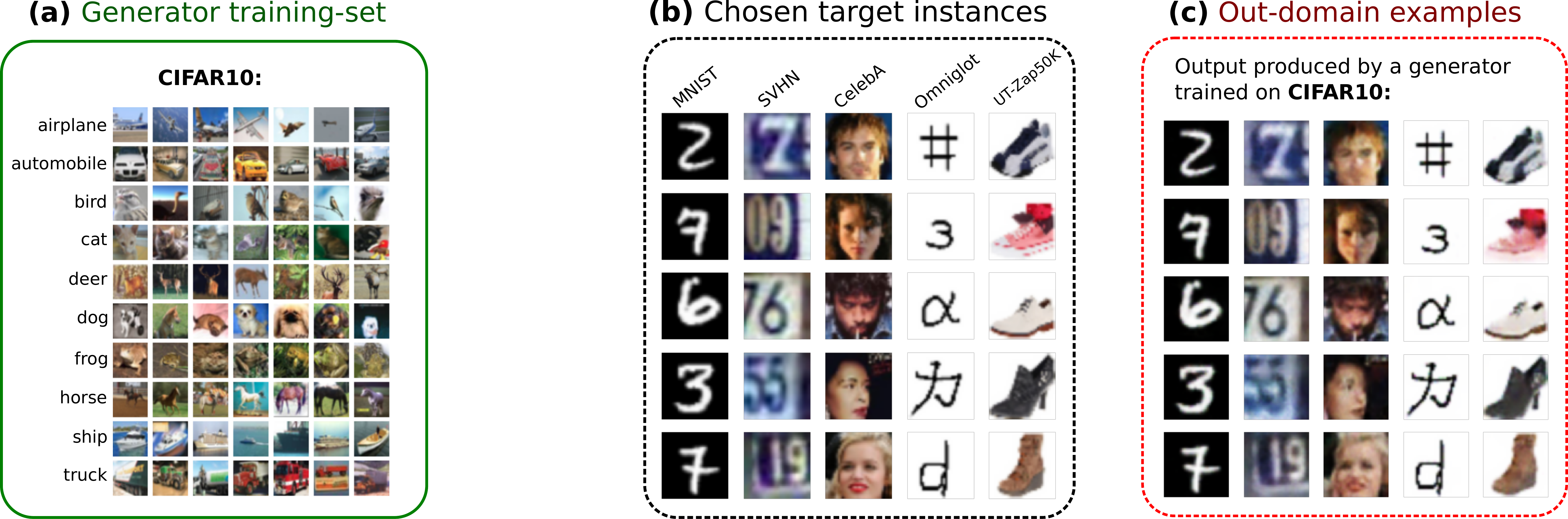}
\caption{\small Target images (panel \textbf{b}) and reproduced out-domain examples (panel \textbf{c})  generated by a \text{DCGAN} generator trained on the \textit{CIFAR10} dataset (panel \textbf{a}). Target images have been randomly chosen from five common image datasets different from \textit{CIFAR10}. The generated out-domain examples are visually close to their respective targets even if there is not intersection between \textit{CIFAR10} and the datasets of the chosen target instances.}
\label{dcgan_mini}
\end{figure*}
\begin{equation}
\label{outdomain_example}
\begin{gathered}
    \G(\zhat)=\xhat 
   \\
    \pdatadist{\xhat \in \mathcal{X}} \leq \epsilon_x  \ \ \text{and} \ \
   \zpriordis{\zhat \notin \zspace} \leq \epsilon_z
\end{gathered}
\end{equation}
where $\zhat$ is the $OLV$ used to generate an out-domain example $\xhat$ and both $\epsilon_x$ and $\epsilon_z$ are negligible probabilities. Hence, the underlying assumption in \eqref{outdomain_example} is that the probability of $\xhat$ of belonging to the set of expected outputs is very low. We will refer to the set of expected outputs of a generator with the term \textit{domain}. The domain of a generator can also be intended as the semantic contents defined by $\pdatadistw$\footnote{For instance, the domain of the \textit{MNIST} dataset is the set of digits representation and the domain of \textit{CelebA} dataset is a set of human faces.}. 
To the purpose of finding a suitable out-domain latent vector $\zhat$, we choose a target instance $\xtarget$ and then, we look for the $\zhat \in \zspace$ such that there is the minimum distance between $\xtarget$ and $\xhat = G(\zhat)$. We refer to this process with the expression \textit{latent search}.
Coherently with \eqref{outdomain_example}, the target instance $\xtarget$ is chosen \emph{ad hoc} to be out of the generator's domain. 
This scenario, is depicted in Figure \ref{dcgan_mini} where a set of out-domain examples (panel \textbf{c}) from a \textit{DCGAN} generator trained on\textbf{} \textit{CIFAR10}  is reported. In this case, $25$ target instances have been randomly chosen from five datasets different from \textit{CIFAR10}.\\
In addition, as required by \eqref{outdomain_example}, an out-domain latent vector $\zhat$ is considered a valid input for the generator $\G$ if it lies in a \emph{dense} region of the latent space. This implies that $\zhat$ must be statistically indistinguishable from a valid latent vector sampled from the latent probability distribution $\zpriordisw$. In the adversarial perspective, this means that a defender is unable to tell apart a valid latent vector from an out-domain latent vector before the generation of $\xhat$.
%
\subsection{Motivating adversarial scenario}
\label{motiv_scen}
The recent success of generative models in the scientific  \cite{particlegan, molgan} and in the entertainment field \cite{animegan, musicgan}, inspired the development of many \textit{GANs} based software applications. These are often in the form of a web service with an interactive interface by which the user provides direct or indirect input to the model\footnote{One example can be found here: \url{https://make.girls.moe}.}. Assuming a white-box access to the generator model, an attacker can find out-domain latent vectors capable of driving the service to produce inadequate contents such as pornographic or offensive material.
The attacker can use these out-domain examples in order to perform a very effective and straightforward defacing attack direct to the generator owner. Indeed, this type of web application and software allows to share and save internal copies of the images created by the users. This scenario resembles a reflected or stored \textit{Cross-site Scripting (XSS)} attack where the attacker is able to arbitrary modify an image in the web page.
The white-box assumption is supported by the observation that often these applications, in order to reduce the server load, run the generative model in the client-side
and additionally pre-trained versions of open-source generator are frequently used.\\\\
We assume that the owner (referred as defender) performs a validation process on $\zhat$ before the calculation of $\xhat = G(\zhat)$. This validation can be intended as a function $\upsilon : \zspace \rightarrow \{0,\ 1\}$. Therefore, the defender accepts to calculate $\xhat$ if and only if $\upsilon(\zhat)$ is equal to $1$. In our attack scenario, this function $\upsilon$ is represented by a distributional hypothesis test. The null hypothesis ($H_0$) is that the vector $\zhat$ is sampled from $\zpriordisw$. Thus, given a test statistic $t$ and for fixed \emph{type I error} $\alpha$, the decision rule can be formalized as follows:
\begin{equation}
    \label{dec_rule}
    Pr(T \geq t|H_0) \geq \alpha \Rightarrow \upsilon(\zhat) = 1
\end{equation}
where $T$ is the distribution of the test statistics under $H_0$ and $Pr(T \geq t|H_0)$ corresponds to the classic  $p$-value of confirmatory data analysis. The same scenario can be easily extended to conditional generators. In that case, we assume that the defender is able to arbitrary choose and fix a data class $y \in Y$. The attacker aims at finding a suitable out-domain latent vector for the conditioned generator $G(\cdot | Y=y)$.
\section{Methodology}
\label{methodology}
As mentioned in Section \ref{ode}, out-domain examples can be found by looking for the closest representation of an arbitrary chosen target instance in the data space defined by $G$. Actually, by leveraging the differentiable nature of the generator and the structure of a well formed latent-data mapping \cite{dcgan}, we can transform this searching problem in an efficient optimization process as follows:
\begin{equation}
\label{latentsearch}
\begin{gathered}
    \loss (\xtarget, \z) = \distfun(\xtarget,\ G(\z)) + \rho(\z) \\
    \zhat = \operatorname*{argmin}_{\z \in \zspace} \loss(\xtarget, z)
\end{gathered}
\end{equation}
Where $\xtarget$ is a given target instance, $\distfun$ is a distance function and $\rho$ is a penalty term applied to $\z$. The purpose of a penalty term is to force the solution $\zhat$ to be consistent with $\zpriordisw$. More precisely, $\rho$ is defined as the weighted sum of the squared difference of the first $k$ sample moments of $z$ and the theoretical moments of a random variable $\zprior \sim \zpriordisw$.  
\begin{equation}
    \label{moments_penalties}
    \rho(z) = \sum_{i=1}^{k} \omega_i \| \mu_\zprior(i) - \Tilde{\mu}_z(i) \|^2_2
\end{equation}
Where $\mu_Z(i)$ is the $i^{th}$ moment of $Z$ and $\Tilde{\mu}_z(i)$ is the $i^{th}$ sample moment of the latent vector $z$. The parameter $\omega_i$ is the weight assigned to the $i^{th}$ moment difference.
In the case of conditional generators, the searching process is performed by fixing a class $y$ as input of the generator function. This implies that the optimization process acts only on the latent vector $z$ and cannot modify the class representation $y$.
More formally, in the conditional case, the problem can be reformulated as:
\begin{equation}
\label{conditionallatentsearch}
\begin{gathered}
    \loss (\xtarget, \z) = \distfun(\xtarget,\ G(\z,\ y)) + \rho(\z) \\
    \zhat = \operatorname*{argmin}_{\z \in \zspace} \loss(\xtarget, z)
\end{gathered}
\end{equation}
assuming that $y$ is randomly chosen from $Y$ by the defender.\\\\
Starting from a random initialization of $\z$, say $\zhatbase$ obtained by sampling from $\zpriordisw$, we iteratively update the current latent vector according to the following rule:
\begin{equation}
    \label{gradient}
    z_{i+1} = z_i + \eta \nabla \loss(\xtarget, z_i), \quad i = 1,\dots,N  
\end{equation}
where $\eta$ is the learning rate.
At each iteration of the optimization process, the distance function $d$ is computed between the target $\xtarget$ and $G(\z_i) = \xhat_i$. We tested and compared two distance functions: the mean squared error (\textit{MSE}) and the cross entropy (\textit{XE}). 
It is noteworthy that in the cross entropy case, the \textit{softmax} function is used in order to ensure the unitary sum in both $\xtarget$ and $G(z)$. However, given its not-bijectivity, we force the comparison between the target and generated image to be scale invariant.
Nonetheless, although this approach diverts from the original objective of founding the closest codomain instance to $\xtarget$, \textit{XE} is able to provide a very good approximation (at least in the visual form) of $\xtarget$ with fewer training iterations than \textit{MSE}.\\
The penalty term is used to ensure the \emph{indistinguishability} of the out-domain vector from a trusted input. The main objective is to find a $\zhat$ such that the probability of $\upsilon(\zhat) = 1$ is maximized. This can be obtained by forcing the out-domain latent vector to have moments equal to those of a random variable distributed as $\zpriordisw$. Indeed, in probability theory, Moment Generating Functions (MGFs) have great practical relevance not only because they can be used to easily derive moments, but also because they uniquely identify a probability distribution function, a feature that makes them a handy tool to solve several problems.
\input{tables/DCGAN_ALL.tex}
The MGF (if it exists) can be seen as an alternative specification useful to characterize a random variable. On one hand, the MGF can be used to compute the $n^{th}$ moment of a distribution as the $n^{th}$ derivative of the MGF evaluated in 0
On the other hand, a MGF is useful to compare two different random variables, studying their behavior under limit conditions. Given a random variable $X$, its MGF is defined as the expected value of $e^{tX}$:
\begin{equation}\label{mgf}
    M_X(t) = \mathbb{E}(e^{tX}), \quad t \in \mathbb{R}
\end{equation}
If \eqref{mgf} holds, then the $n^{th}$ moment of $X$, denoted by $\mu_X(n)$, exists and it is finite for any  $n \in \mathbb{N}$:
\begin{equation}
    \mu_X(n) = \mathbb{E}(X^n) = \frac{\partial^n M_X(t)}{\partial t^n}\bigg\rvert_{t = 0}
\end{equation}
\section{Results}
\label{results_exp}
In our experiments, we tested and compared two common prior distributions, i.e., the standard normal and the continuous uniform distribution in $[-1,1]$. Given the constraint imposed by the latter, we perform a hard clipping on $z$ values in order to force the latent vector to lie in the allowed hypercube. As proposed in \cite{precise_recovery}, we tested the \textit{stochastic clipping} method but results showed no substantial improvement.\\
We did not apply any clipping method for the normal prior distribution. Empirically, it has been observed that the penalty on the moments is sufficient to guarantee that $z$ assumes values in an acceptable range.
The quality of the out-domain examples is evaluated on different \textit{DCGAN} generators and on  conditional generators trained within an \textit{ACGAN} framework. For the sake of exposition, we will refer to each trained generator with the following compact notation: 
\begin{center}
\textit{[Architecture]}-\textit{[Training-dataset]}-\textit{[Latent\_prior]}-\textit{[Latent\_space\_dimension]}
\end{center}
In particular, a generator is trained for any combination of the followings:
\begin{itemize}
\itemsep0em 
\item \textbf{Architecture:} \textit{DCGAN}, \textit{ACGAN}
\item \textbf{Training Dataset:} \textit{CIFAR10} \cite{cifar10}, \textit{SVHN} \cite{svhn} and a simple variation  of \textit{MNIST} \cite{mnist}, called \textit{ColorMNIST}
    \item \textbf{Latent space dimension:} $\{100,\ 256,\ 512\}$
    \item \textbf{Latent prior distribution:} $N(0, 1)$ and $U[-1,\ 1]$
\end{itemize}
For instance, \cn{DCGAN}{CIFAR10}{Normal}{100} defines a \textit{DCGAN} generator trained on \textit{CIFAR10} with a normal latent prior distribution and latent space dimension equal to $100$. The \textit{ColorMNIST} dataset is obtained by applying a random background color to the original \textit{MNIST}. The reason of that modification is to offer to the generator the chance of representing a larger set of outputs by letting the generator learn a larger number of \textit{RGB} triplets, while keeping virtually unaltered the complexity of \textit{MNIST}. All the generators and discriminators' architectures as well as the hyper-parameters and the training process are the same proposed in \cite{dcgan}. We tested three values of the latent space dimension that are commonly used in literature.\\
The validation process of the out-domain vectors is performed by fixing $\alpha = 0.05$, $k = 4$ for the normal prior and $k = 6$ for the uniform prior. We performed three different distributional tests, i.e., Kolmogorov-Smirnov, Shapiro-Wilk and Anderson-Darling \cite{shapiro1990test}. 
Results showed that, given the penalty term $\rho$, all the distributional tests bring to the same decision. The following results refer to the Anderson-Darling test \cite{anderson1954test}, which was finally chosen since its test statistics is based on the \emph{Cumulative Distribution Function} \textit{CDF}\cite{ross2014first} and, compared to other tests, it places more weight on the values in the tails of the distribution.\\
We defined a test-set of target instances to the purpose of evaluating the capability of different generators to reproduce out-domain examples. This test-set contains randomly chosen instances from four image datasets i.e, \textit{Omniglot} \cite{omniglot},  \textit{CelebA} \cite{celeba}, \textit{UT-Zap50K} \cite{shoes} and \textit{Tiny ImageNet} \cite{tinyimagenet}.
A random sample of $32$ images is selected for each dataset for a total of $128$ target instances. In the case of \textit{Tiny ImageNet}, the images are sampled from classes which are different from those of \textit{CIFAR10}.
All images are forced to share the same dimension of $32 \times 32$ pixels and to be normalized in the interval $[0, 1]$. To simplify the understanding of the results, the target distance function used for all the experiments is the \emph{Mean Squared Error} (\textit{MSE}).
The average \emph{Mean Squared Error} \textit{MSE}
and the percentage of successfully passed statistical tests are computed on the test-set and used as main evaluation scores. In the latent search process, the \textit{Adam} \cite{adam} optimizer is used with a learning rate equal to $0.01$. All the experiments are performed using the \textit{TensorFlow} framework \cite{tensorflow}. The most relevant codes used for the present  work along with an interactive proof of concept are available on: \url{https://github.com/pasquini-dario/OutDomainExamples}.
\begin{figure*}
\centering
\includegraphics[scale=0.32]{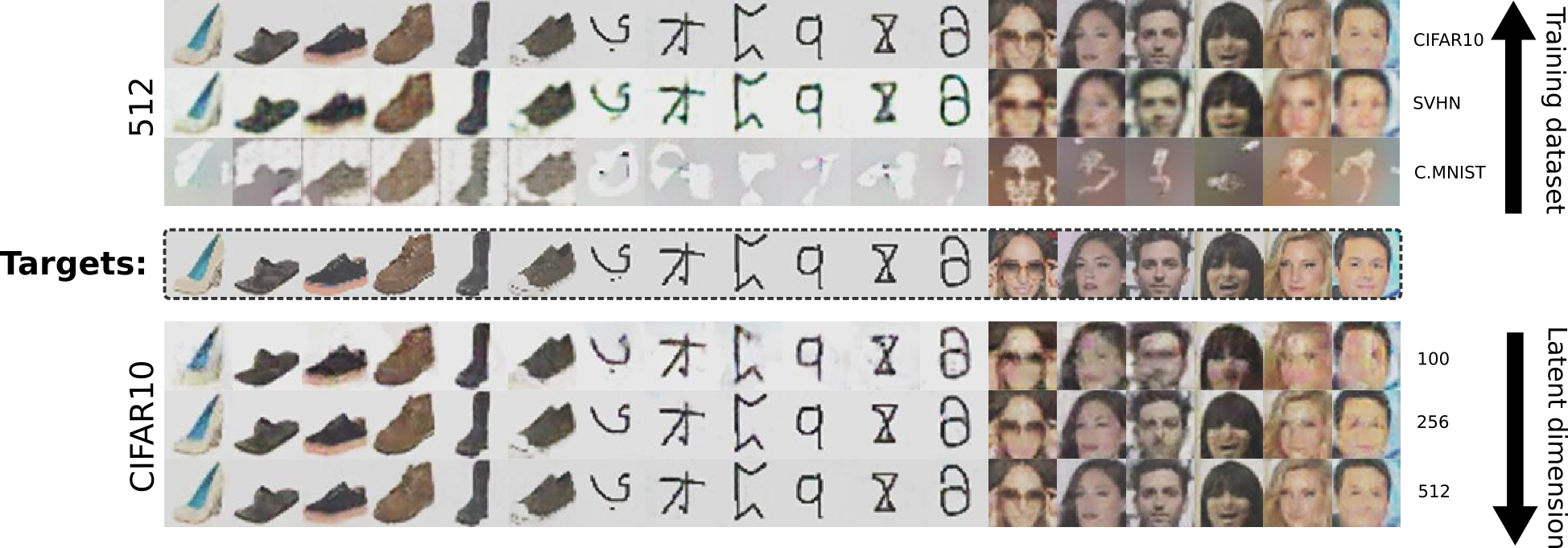}
\caption{Out-domain examples for a set of \textit{DCGAN} architectures trained with uniform latent prior. The central row shows several different randomly chosen targets from the test-set. The upper panel shows the variability in the out-domain generation when the latent space dimension is fixed to $512$ and the training dataset of the generator varies. The lower panel shows the variability in the out-domain generation when the training dataset of the generator is fixed to \textit{CIFAR10} and the latent space dimension varies.}
\label{dcgan_uniform_cifar10_latent_size_sample_comparison}
\end{figure*}
\subsection{DCGAN}
\label{dcgan_exp}
\begin{figure*}[ht]
\begin{minipage}[t]{0.38\linewidth} 
    \centering
    \includegraphics[scale=0.40]{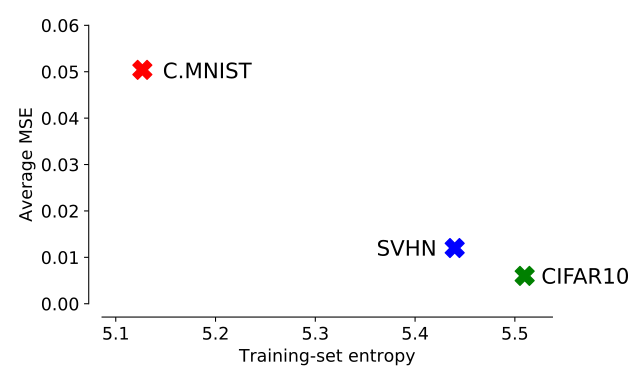}
    \caption{Average \textit{MSE} compared to the estimated entropy of each training set.}
    \label{entropy_mse}
\end{minipage}  
\hspace{0.1cm}
\begin{minipage}[t]{0.62\linewidth}
    \centering
    \includegraphics[scale=0.36]{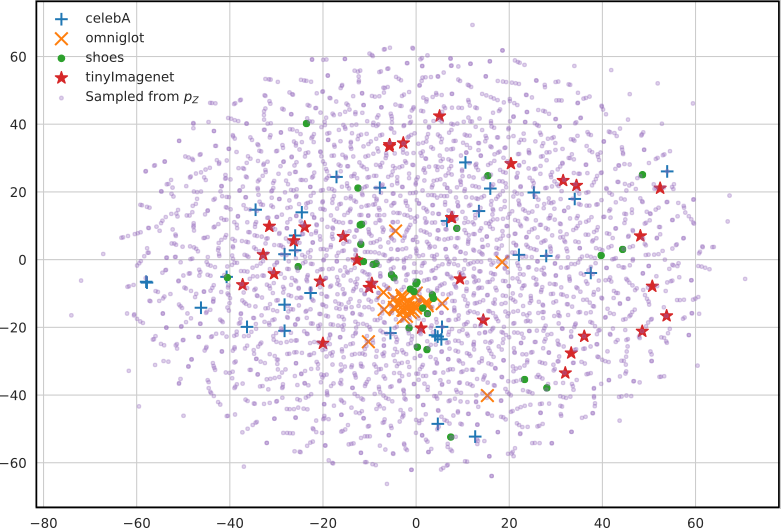}
    \caption{\small Two-dimensional representation of out-domain latent vectors for the \cn{DCGAN}{CIFAR10}{Uniform}{100}. Out-domain latent vectors with target instances coming from the same dataset are represented with the same marker.}
    \label{tsne}
\end{minipage}      
\end{figure*} 
Table \ref{dcgan_allresulttable} shows the results related to the \textit{DCGAN} generators. 
\begin{figure*}
    \centering
    \includegraphics[scale=0.44]{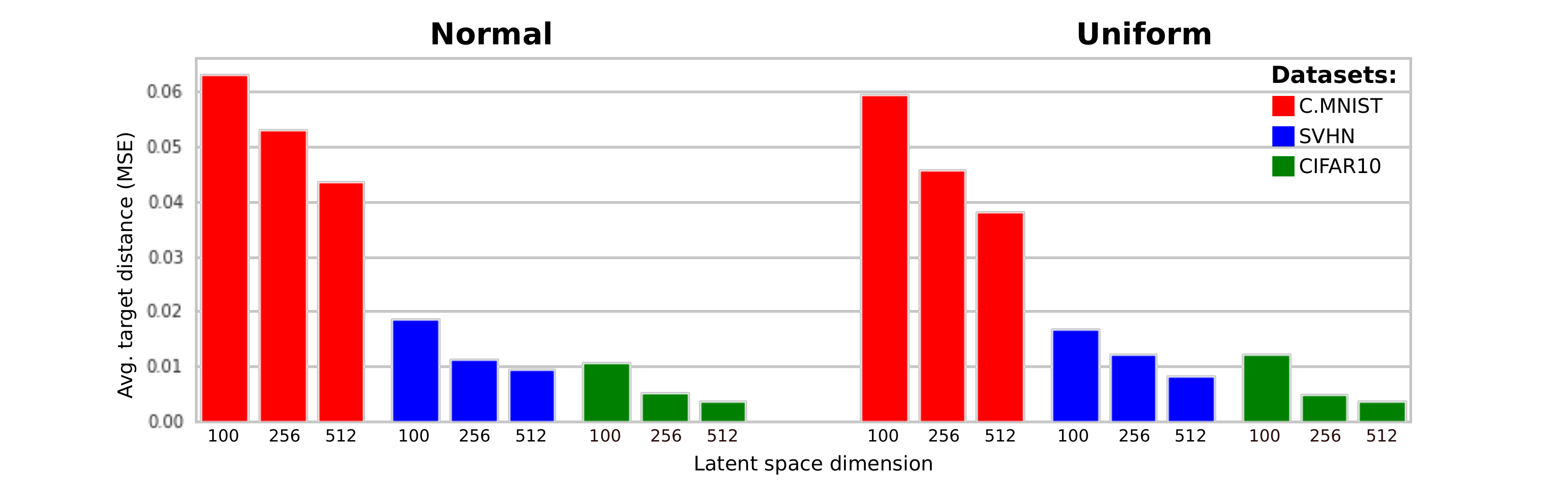}
    \quad
    \caption{ \small Average \textit{MSE} for all the \textit{DCGAN} generators trained on all combinations of dataset, latent space dimension and latent prior distribution.}
    \label{dcgan_all_mse}
\end{figure*}
All the forged \odlv pass successfully the distributional test, regardless of the chosen prior distribution. Several checks on the biases of the \odlv have been carried out providing pretty good results. In particular, the estimation of the odd moments for the normal is precise; the second and the fourth moment are instead slightly overestimated and underestimated, respectively. For the uniform prior, the bias for the second and the sixth moment is slightly positive and for the remainings the estimation is precise, with a quite large variance in the estimation of the third moment. No evident patterns are worth to notice when looking at the training dataset or at changes of the latent space dimension for both the prior distributions. These results are due to the fact that the penalty term $\rho$ strongly constraints the values of the latent vectors in a well defined range. It is worth to notice that relaxing the moments penalty during the optimization process \eqref{latentsearch} would reduce further the \textit{MSE}. In contrast to in-domain inversion \cite{creswell2018inverting}, we can state that out-domain examples do not take any significant advantage from latent vectors statistically close to those used during the training process.\\
Figure \ref{dcgan_uniform_cifar10_latent_size_sample_comparison} shows a set of target instances and related out-domain examples for a total of six generators. The upper panel reports the out-domain examples produced by three generators trained on different training sets but with same latent space dimension and prior. When the training set of the generator is \textit{ColorMNIST}, the method fails in finding suitable \odlv capable of reproducing the target images. For the other two, the generator is able to provide a valid reconstruction for all the targets. The failure of \textit{ColorMNIST} may be connected to the fact that it is less \emph{heterogeneous} with respect to \textit{SVHN} and  \textit{CIFAR10}. By heterogeneity we intend the actual number of different pixels which are necessary in order to reproduce the same heterogeneity of the whole set  (i.e., the entropy). It is reasonable to expect that the larger the variety of images in the training set, the larger will be the set of potential out-domain examples reproduced by the generator. As an estimator of that variety, we computed the Shannon entropy\cite{jost2006entropy} for a sample of $2^{10}$ images from each training set. 
%
%
\input{images/Interpolation/interpolation.tex}
Results are depicted in Figure \ref{entropy_mse} and show that there is a strong dependence between the average \textit{MSE} (i.e., the reconstruction error) and the entropy of the training set.\\ Figure \ref{tsne} depicts a two-dimensional projection of a set of out-domain latent vectors and latent vectors directly sampled by $p_Z$. This representation is obtained by applying the dimension reduction algorithm called \emph{t-distributed Stochastic Neighborhood Embedding} (\textit{t-SNE}) \cite{maaten2008visualizing} on vectors of size $100$. It is possible to note how the out-domain latent vectors tend to be uniformly distributed in the space. In the case of the \textit{Omniglot} dataset, the \odlv tend to cluster in a specific region and this may be due to its intrinsic homogeneity.\\
Even if the entropy is a sort of predictor of the success of our method, it is still possible, given a target image, a latent prior and a training set, to enhance the quality of the generated image by increasing the dimension of the latent space. As a matter of fact, we can observe, by looking at the lower panel of Figure \ref{dcgan_uniform_cifar10_latent_size_sample_comparison}, that an increase of the latent space dimension makes the generated image more similar to the target one. An additional motivation can be that the latent space acts as an information bottleneck for the target instance during the latent search process
All these possibilities are evaluated in terms of \textit{MSE}. Figure \ref{dcgan_all_mse} shows the average \textit{MSE} for each latent space dimension, latent prior and training set confirming that the more is the entropy of the training set, the higher the probability of success in the generation and, at the same time, the larger the dimension of the latent space, the higher the quality of the reconstruction. Instead, there is no relevant difference in the quality of the out-domain examples when the latent prior distribution varies.\\
Figure \ref{progan_linterpolation} shows three examples of linear interpolation between latent vectors \cite{dcgan}. The first row depicts a smooth and semantic meaningful transaction between two random vectors sampled from $\zpriordisw$, referred as \textit{in-domain} latent vectors. By semantic meaningful transaction, we mean that each image between the two interpolation points remains coherent with $\pdatadistw$. The second row depicts the interpolation between an in-domain vector and an out-domain vector. In contrast with the first case, the transaction is unbalanced and not particularly smooth. From the sequence, it can be noticed that the semantic valid attributes of the starting image, i.e. the black of the hair and the reflection on the forehead, are deformed to recreate the final \textit{MNIST} digit. 
The last row shows the extreme case of interpolation between two out-domain latent vectors. In this case, all the intermediate data instances never cross the in-domain set.
%
%
%
\input{tables/ACGAN_ALL.tex}
\begin{figure*}
\begin{minipage}[t]{0.5\linewidth}
    \centering
    \includegraphics[scale=0.2]{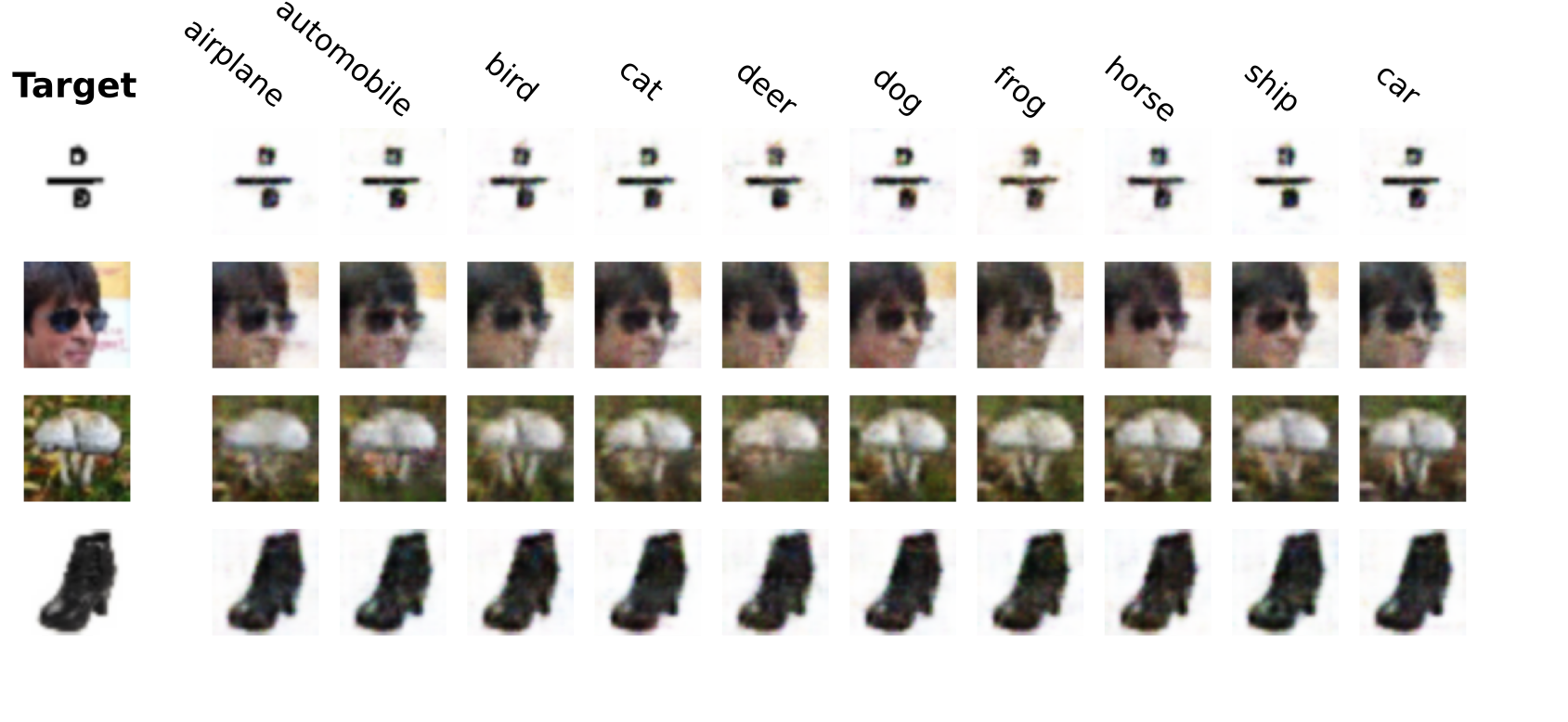}
    \caption{ \small Comparison between out-domain examples produced by an \cn{ACGAN}{CIFAR10}{Normal}{512}. Each (but the first) column  depicts the out-domain example produced by the generator, conditionally to each class.}
    \label{ACGAN_CIFAR10_512_uniform}
\end{minipage}
\hspace{0.1cm}
\begin{minipage}[t]{0.5\linewidth} 
\includegraphics[scale=0.32]{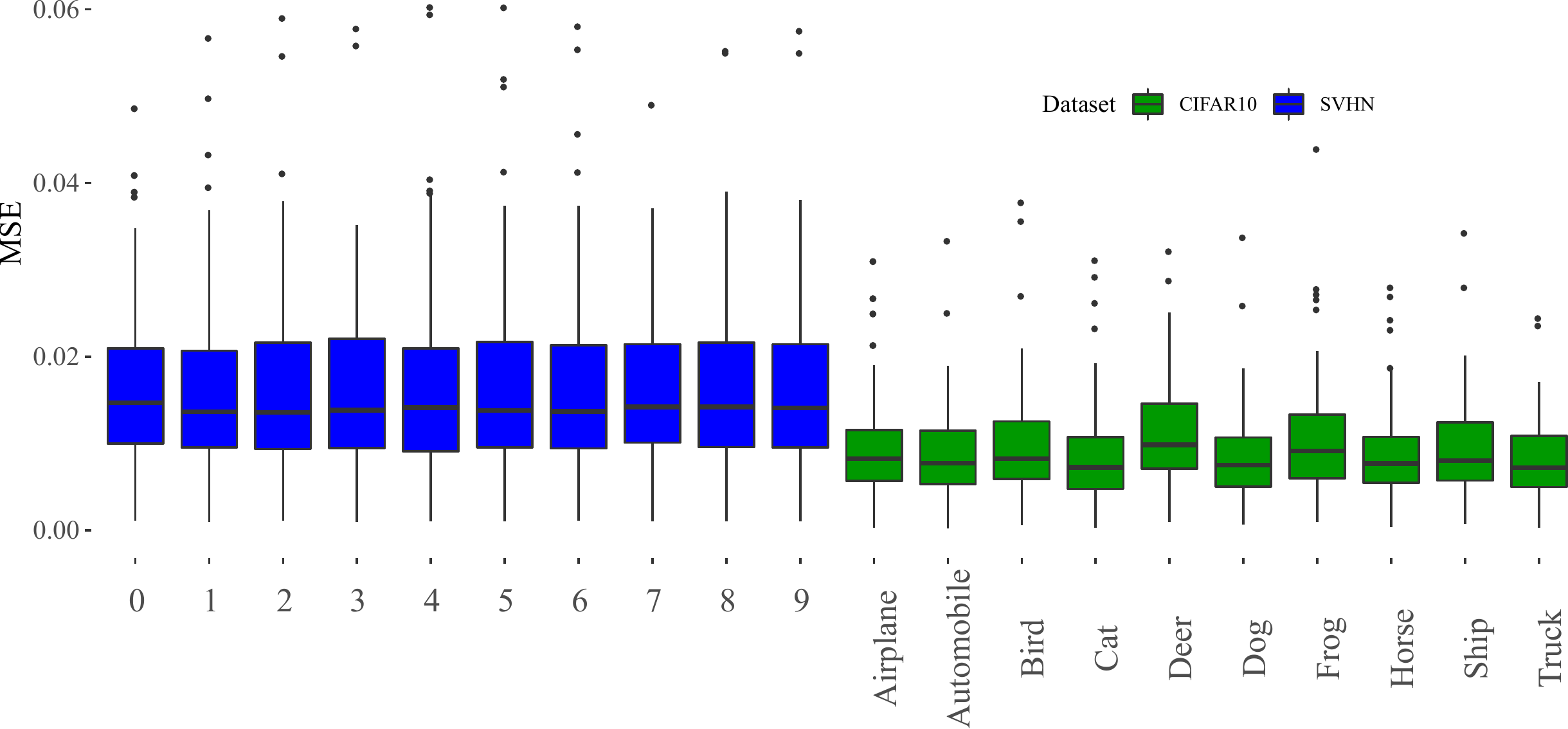}
\caption{\textit{MSE} distribution conditionally to each class for \textit{CIFAR10} and \textit{SVHN} datasets.}
\label{acgan_mse_per_class.png}
\end{minipage}        
\end{figure*} 
\subsection{ACGAN}
\label{acgan_exp}
Conditional generators are trained to the purpose of enforcing their outputs to be part of a meaningful, from the semantic viewpoint, data class. Typically, this implies a better global coherence and quality  in the definition of the generator's data space \cite{acgan}. This is especially true for models trained with the \textit{ACGAN} framework in which the generators are encouraged to produce images that are correctly classified from the discriminator as genuine and belonging to its class. The experiments described below aim at finding out if the conditional extension is sufficient to the purpose of preventing the generation of out-domain examples.
We trained different \textit{ACGAN} generators using the same set of parameters reported in Section \ref{dcgan_exp}. 
Also the architecture used for the generator and the discriminator is the same used for the previous \textit{DCGAN} experiments\footnote{The only difference is the number of neurons in the generator's input layer and in the discriminator's output layer due to the conditional setup}. Training hyper-parameters are the same proposed in \cite{acgan}.
As aforementioned, in the conditional setup, the hypothesis is that the class $y$ is randomly chosen by the defender and the attacker can not modify its representation during the latent search process.
Tests and validation are performed as in Section \ref{dcgan_exp} but they are evaluated conditionally to each class $y$ in the generators' classes set. All the tested training sets are composed by $10$ classes. In this case, the \textit{MSE} is calculated as the average over the classes.\\
%
%
%
%
We are able to find an out-domain example for each image in the test-set, conditionally to each class. We do not report the results when the training set is \textit{ColorMNIST},  since it already failed in the less severe \textit{DCGAN} experiment. As an example, Figure \ref{ACGAN_CIFAR10_512_uniform} shows the generated out-domain examples, conditionally to each class of \textit{CIFAR10}, for four randomly chosen target instances in the test-set. It can be noticed that the class has no relevant impact on the quality of the out-domain examples: the attack succeeds regardless of the class. The same happens when attacking the generators trained on \textit{SVHN}.
Results in terms of \textit{MSE} are summarized in Table \ref{acgan_allresulttable}. It is possible to observe that the average \textit{MSE} is uniformly larger compared to the \textit{DCGAN} experiments due to the conditional setup. In Figure \ref{acgan_mse_per_class.png}, we also report the distribution of the \textit{MSE}, conditionally to each class, for each training set.
No specific patterns are registered: the \textit{MSE} distribution is approximately the same for each class and training set. However, there is a slight variability for \textit{CIFAR10} given the larger heterogeneity among its classes.
The validation of the out-domain latent vectors is the same described in Section \ref{dcgan_exp}. Also in this case, all the latent vectors pass successfully the Anderson-Darling test. Moments distributions for each dataset, latent prior and latent space dimension are also checked. No relevant difference has been observed with respect to the not-conditional generators experiments.
\section{Conclusion and further developments}
\label{concl_fd}
We showed how to forge suitable adversarial inputs capable of driving a trained generator to produce arbitrary data instances in output. This is possible for both conditional and not-conditional generators. Additionally, we showed that an adversarial input can be shaped in order to be statistically indistinguishable from the set of trusted inputs. We also showed that the success of our method strongly depends on two main factors: the heterogeneity of the set on which the generator is trained and the latent space dimension.\\ 
In additional experiments we found a set of generators showing a greater resilience to the generation of out-domain examples. In particular, the \textit{Non-saturating GAN}s with \textit{ResNet} architecture analyzed in \cite{ganland} shows an inherent difficulty to produce out-domain examples even when the generator is trained on high entropic datasets such as \textit{CIFAR10}. We conjecture that this property is strongly related to the generator architecture.\\
In the described adversarial scenario, we supposed that an aware defender can just test the validity of the model's input in order to evaluate the genuineness of the latent vectors. However, it is possible to imagine a more powerful defender able to verify the generator's output in order to spot unexpected generation.\\ 
As future directions of activity, we expect to \emph{i)}  investigate the generation of out-domain examples for other \textit{GAN} architectures; \emph{ii)} study the generation of out-domain examples in contexts other than those of images; \emph{iii)} investigate the possibility of training an arbitrary complex generator which is resilient to the generation of out-domain examples; \emph{iv)} evaluate the possibility of extending the attack to a black-box scenario using an approach inspired by \cite{blackbox}.\\


%

%
%
\bibliographystyle{abbrv}
\bibliography{main_1}
\end{document}

%% file: tables/DCGAN_ALL.tex
\begin{table*}[ht]
\footnotesize
\caption{\small Results concerning the out-domain generation process on the test-set for all the \textit{DCGAN} generators. The column \textbf{Test Succ.} reports the percentage of out-domain latent vectors which successfully passed statistical tests. The column \textit{Avg MSE*} reports the average \textit{MSE} in the case of complete relaxation of the penalty term $\rho$.}
\centering
\begin{tabular}{lr|rrr||rrr|}
\multicolumn{2}{c}{}
&\multicolumn{3}{|c||}{(a) \textbf{Normal Latent distribution}}
& \multicolumn{3}{c|}{(b) \textbf{Uniform Latent distribution}} \\
\toprule
   Dataset &  $Z$ dim. &  \textbf{Avg MSE} & \textbf{Test Succ.} & Avg MSE* &  \textbf{Avg MSE} & \textbf{Test Succ.} & Avg MSE* \\ \midrule
    CIFAR10 &          100 &  0.010646  & 100\% & 0.008881 & 0.012131  & 100\% & 0.009354\\
    CIFAR10 &          256 &  0.005094  & 100\% & 0.003902 &  0.012131  & 100\% & 0.009354\\
    CIFAR10 &          512 &  0.003693  & 100\% & 0.002710 &  0.003603  & 100\% & 0.002826\\ \midrule
       SVHN &          100 &  0.018569  & 100\% & 0.011541 & 0.016730 & 100\% & 0.011359   \\
       SVHN &          256 &  0.011374  & 100\% & 0.006970 &  0.012121 & 100\%  & 0.007213  \\ 
       SVHN &          512 &  0.009474  & 100\% & 0.005314 &  0.008323  & 100\% & 0.005594 \\ \midrule
 C.MNIST &          100 &  0.063097  & 100\% & 0.040453  &  0.059342  & 100\% & 0.037457 \\
 C.MNIST &          256 &  0.052926  & 100\% & 0.029160 &  0.045851  & 100\% & 0.027178\\
 C.MNIST &          512 &  0.043685  & 100\% & 0.025855 &  0.037946 & 99\%  & 0.022415\\
\bottomrule
\end{tabular}
\label{dcgan_allresulttable}
\end{table*}

%% file: images/Interpolation/interpolation.tex
\begin{figure*}[ht]
    \footnotesize
    \centering
    \textbf{a)} In-domain to In-domain\\
    \includegraphics[scale=.20]{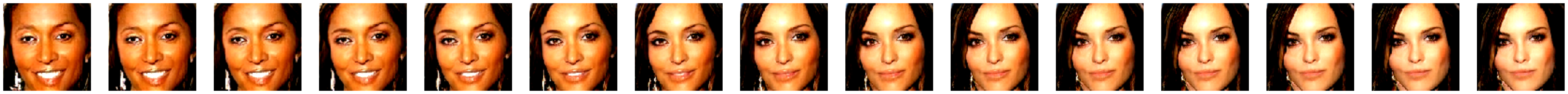}
    
    \textbf{b)} In-domain to Out-domain\\
    \includegraphics[scale=.20]{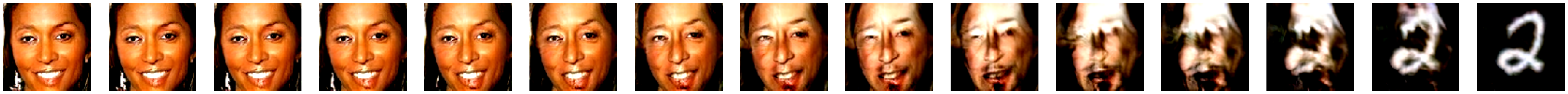}

    \textbf{c)} Out-domain to Out-domain\\
    \includegraphics[scale=.20]{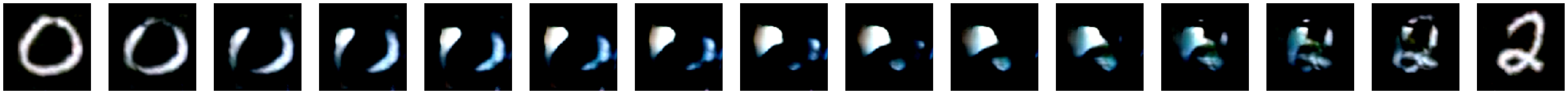}
    \caption{Three examples of linear interpolation between two latent vectors for a \textit{ProGAN} trained on the \textit{CelebA} dataset. The row (a) depicts the interpolation process between two randomly chosen latent vectors. Row (b) depicts interpolation from a randomly chosen latent vector and an out-domain latent vector. Row (c) depicts interpolation between two out-domain latent vectors.}
    \label{progan_linterpolation}
\end{figure*}

%% file: tables/ACGAN_ALL.tex
\begin{table*}[tp]
\footnotesize
\caption{\small Results for any \textit{ACGAN} generator. Scores are obtained as the average over test-set results for each of the ten classes present in the generator training-set.}
\centering
\begin{tabular}{lr|rr||rr|}
\multicolumn{2}{c}{}
&\multicolumn{2}{|c||}{(a) \textbf{Normal Latent distribution}}
& \multicolumn{2}{c|}{(b) \textbf{Uniform Latent distribution}} \\
\toprule
   Dataset &  $Z$ dim. &  \textbf{Avg MSE}  & \textbf{Test Succ.} &  \textbf{Avg MSE}  & \textbf{Test Succ.}  \\ \midrule
 CIFAR10 &          100 &  0.023457 &          100\% &  0.019615 &  100\%\\
 CIFAR10 &          256 &  0.013144 &          100\% &  0.009547 &          100\%\\
 CIFAR10 &          512 &  0.009075 &          100\% &  0.005944 &           99\%\\ \midrule
    SVHN &          100 &  0.026686 &          100\% &  0.024732 &          100\% \\
    SVHN &          256 &  0.016879 &          100\% &  0.016268 &          100\%\\
    SVHN &          512 &  0.016291 &          100\% &  0.013707 &           99\%\\
\bottomrule
\end{tabular}
\label{acgan_allresulttable}
\end{table*}